\documentclass{article}
\usepackage{spconf}
\usepackage{amsmath}
\usepackage{graphicx}
\usepackage{booktabs}
\usepackage[T1]{fontenc}
\usepackage[utf8]{inputenc}
\usepackage{tipa}
\usepackage[normalem]{ulem}
\usepackage{microtype}

\makeatletter
\long\def\@makecaption#1#2{%
  \vskip\abovecaptionskip
  \setbox\@tempboxa\hbox{#1. #2}%
  \ifdim\wd\@tempboxa>\hsize #1. #2\par
  \else \hbox to\hsize{\hfil\box\@tempboxa\hfil}%
  \fi}
\makeatother
\newcommand{\trans}[1]{\uline{#1}}
\usepackage[nopostdot,nonumberlist,acronym]{glossaries}
\usepackage{cleveref}
\crefname{table}{Table}{Tables}
\crefname{figure}{Figure}{Figures}
\crefformat{section}{\S#2#1#3}
\crefmultiformat{section}{\S#2#1#3}{ and \S#2#1#3}{, \S#2#1#3}{ and \S#2#1#3}
\crefname{equation}{Eq.}{Eqs.}

\makeglossaries
\newacronym{llm}{LLM}{large language model}
\newacronym{alm}{ALM}{audio language model}
\newacronym{ipa}{IPA}{International Phonetic Alphabet}
\newacronym{rsa}{RSA}{representational similarity analysis}
\newacronym{ssl}{SSL}{self-supervised learning}
\newacronym{asr}{ASR}{automatic speech recognition}

\title{Do Audio Language Models Hear and Read Distinctive Features Alike?}

\name{Yuanhao Chen \qquad Peter Chin\thanks{This research was funded by the Defense Advanced Research Projects Agency (DARPA), under contract W912CG23C0031.}}
\address{Thayer School of Engineering, Dartmouth College, Hanover, NH, USA \\
  \texttt{yc.th@dartmouth.edu} \qquad \texttt{pc@dartmouth.edu}}

\begin{document}
\maketitle

\begin{abstract}
Audio language models pass speech and text through a single decoder.
We ask whether that decoder represents a distinctive feature in the same direction when a phoneme is heard and when it is read.
For minimal pairs of phonemes differing in one feature, we take the offset between the two members' mean representations.
Averaging those offsets gives a direction for each stream, and we measure the cosine between the two.
Because the two streams already agree about arbitrary phoneme pairs, we compare every measure against a reference built from random pairings rather than against zero.
We apply this to 6 models, 7 features and 15 languages from 11 families.
Only voicing in the two Qwen2.5-Omni models exceeds that reference after correction for multiple testing, and the reference varies by a factor of seven between models.
In three of the six models, voicing has one direction in audio across the 14 languages with enough minimal pairs to measure it, and every language pair agrees in two of them.
The model family, not the model size, predicts which stream represents a feature.
\end{abstract}

\begin{keywords}
audio language models, phonological features, cross-modal representation, multilingual speech
\end{keywords}

\section{Introduction}
\label{sec:introduction}

An \gls{alm} receives speech through an audio encoder and text through a tokeniser, and both streams pass through one decoder.
We study whether that decoder represents a distinctive feature as one direction in both streams, or keeps a separate representation for each.

Work on self-supervised speech models suggests that distinctive features are encoded as directions.
Choi et al. \cite{choiSelfsupervisedSpeechModels2026} show that distinctive features behave additively in such models, so that the offset between [d] and [t], added to [p], gives [b].
Phonetic information in these models also has a characteristic depth profile \cite{pasadLayerWiseAnalysisSelfSupervised2021}, and a multilingual one learns latent units shared across languages \cite{conneauUnsupervisedCrosslingualRepresentation2020}.
ALAS \cite{mousaviALASAutomaticLatent2026} runs an \gls{alm} on audio and on the transcript of the same audio, scoring per layer how well an audio frame's hidden state identifies the text token it corresponds to.
That measures the temporal binding between the two streams rather than the geometry of any one feature.

We take minimal pairs of phonemes differing in one distinctive feature, in 15 languages from 11 families, and compare the direction of their offsets when the phonemes are heard and when they are read.
Every measure has a reference drawn from random pairings rather than a comparison against zero.
After correction for multiple testing, only voicing in the two Qwen2.5-Omni models exceeds that reference, which itself differs more between models than any feature effect does.
Three of the six models use one voicing direction in audio for every language in which voicing can be measured.
Which stream represents a feature is predicted by the model family and not by the model size.

\section{Method}
\label{sec:method}

\subsection{Corpus, languages and models}
\label{sec:corpus}

We draw phonemes from the Language Documentation Reference Corpus \cite{paschenBuildingTimeAlignedCrossLinguistic2020}.
It provides time-aligned segments of spoken narrative, labelled with a broad transcription in the \gls{ipa}.
We keep the rows that annotate speech and discard pauses and disfluency markers.
Its transcription needs some canonicalisation before a feature table reads it (e.g., an ASCII \texttt{g} appears across 26,272 occurrences where the \gls{ipa} voiced velar stop is expected).

A language enters the study on three conditions:
(1) Its audio must be distributed with the corpus.
(2) Its segments must be linked to recordings.
(3) At least one of the 7 features in \cref{tab:languages} must have three or more minimal pairs among phoneme types occurring at least 100 times, though two pairs suffice to measure a feature.
Forty of the 47 languages qualify.
We rank those by how many features meet that threshold, breaking ties in favour of a less represented family, and take the first 15.
\Cref{tab:languages} lists them, across 11 families.

Phonemes are grouped into the utterance containing them.
A group's span runs from its first phoneme's start to its last phoneme's end, so the audio a model receives contains every phoneme we then locate within it.

We study Qwen2-Audio \cite{chuQwen2AudioTechnicalReport2024}, Qwen2.5-Omni at two sizes \cite{xuQwen25OmniTechnicalReport2025}, Gemma 4 at two sizes \cite{teamGemma4Technical2026}, and Audio-Flamingo 3 \cite{ghoshAudioFlamingo32025}.

\begin{table}[t]
\centering
\small
\setlength{\tabcolsep}{3pt}
\caption{The language sample. \emph{Features} are those with at least two minimal pairs among phoneme types occurring at least 100 times: \emph{voi}~voicing, \emph{hi}~height, \emph{long}~length, \emph{back}~backness, \emph{sg}~spread glottis, \emph{nas}~nasality, \emph{cg}~constricted glottis.}
\label{tab:languages}
\begin{tabular}{@{}lll@{}}
\hline
Language & Family & Features \\
\hline
Baïnounk Gubëeher~\cite{doreco-bain1259} & Atlantic-Congo & back, long, voi \\
Ruuli~\cite{doreco-ruul1235} & Atlantic-Congo & back, hi, long, voi \\
\hline
Bora~\cite{doreco-bora1263} & Boran & back, hi, long, sg \\
\hline
Cabécar~\cite{doreco-cabe1245} & Chibchan & back, hi, nas, voi \\
\hline
French (Swiss)~\cite{doreco-stan1290} & Indo-European & back, nas, voi \\
\hline
Svan~\cite{doreco-svan1243} & Kartvelian & cg, hi, long, voi \\
\hline
Texistepec Popoluca~\cite{doreco-texi1237} & Mixe-Zoque & back, long, nas, voi \\
\hline
Sanzhi Dargwa~\cite{doreco-sanz1248} & Nakh-Daghestanian & cg, hi, long, voi \\
Tabasaran~\cite{doreco-taba1259} & Nakh-Daghestanian & cg, hi, voi \\
\hline
Anal~\cite{doreco-anal1239} & Sino-Tibetan & long, sg, voi \\
Sadu~\cite{doreco-sadu1234} & Sino-Tibetan & back, hi, sg, voi \\
Sümi~\cite{doreco-sumi1235} & Sino-Tibetan & hi, sg, voi \\
\hline
Evenki~\cite{doreco-even1259} & Tungusic & hi, long, voi \\
\hline
Dolgan~\cite{doreco-dolg1241} & Turkic & back, hi, long, voi \\
\hline
Kamas~\cite{doreco-kama1351} & Uralic & hi, long, voi \\
\hline
\end{tabular}
\end{table}

\subsection{Locating a phoneme in audio and in text}
\label{sec:locating}

\begin{table*}[t]
\centering
\small
\setlength{\tabcolsep}{3pt}
\caption{Median $c$ per model and feature, at the layer where the median over languages is largest. Cells give that median ($\bar{c}$), the 95th percentile of its reference ($c_{95}$), the number of languages with a positive $c$ ($+$), and the number of languages whose $c$ exceeds its largest value against another feature ($\times$). An asterisk marks $q \leq 0.05$ over the 42 tests.}
\label{tab:cross-modal}
\begin{tabular}{@{}lcccccccccccccccc@{}}
\hline
 & \multicolumn{4}{c}{Voicing (14 lang.)} & \multicolumn{4}{c}{Height (11 lang.)} & \multicolumn{4}{c}{Length (10 lang.)} & \multicolumn{4}{c}{Backness (8 lang.)} \\
Model & $\bar{c}$ & $c_{95}$ & $+$ & $\times$ & $\bar{c}$ & $c_{95}$ & $+$ & $\times$ & $\bar{c}$ & $c_{95}$ & $+$ & $\times$ & $\bar{c}$ & $c_{95}$ & $+$ & $\times$ \\
\hline
Qwen2-Audio 7B & +0.54 & +0.75 & 12 & 7 & +0.68 & +0.80 & 8 & 5 & +0.36 & +0.78 & 8 & 3 & +0.10 & +0.78 & 7 & 1 \\
Qwen2.5-Omni 7B & +0.40$^{*}$ & +0.34 & 14 & 14 & +0.27 & +0.35 & 11 & 11 & +0.17 & +0.35 & 10 & 6 & +0.27 & +0.34 & 6 & 5 \\
Qwen2.5-Omni 3B & +0.45$^{*}$ & +0.35 & 14 & 14 & +0.20 & +0.36 & 11 & 11 & +0.10 & +0.35 & 8 & 5 & +0.27 & +0.35 & 7 & 7 \\
\hline
Gemma 4 E4B & +0.07 & +0.11 & 9 & 7 & +0.08 & +0.12 & 8 & 4 & +0.08 & +0.12 & 8 & 3 & +0.10 & +0.13 & 7 & 5 \\
Gemma 4 E2B & +0.11 & +0.12 & 12 & 7 & +0.08 & +0.13 & 8 & 2 & +0.11 & +0.13 & 7 & 4 & +0.07 & +0.14 & 7 & 0 \\
\hline
Audio-Flamingo 3 & +0.18 & +0.20 & 13 & 13 & +0.15 & +0.22 & 11 & 10 & +0.08 & +0.21 & 7 & 2 & +0.16 & +0.22 & 7 & 5 \\
\hline
\end{tabular}
\end{table*}

To map decoder position to time we encode clips of several durations and count the positions holding the model's audio placeholder.
The count grows in proportion to duration, and the ratio gives milliseconds per position $r$.

For the audio stream we cut each recording to a group's span and run one forward pass, keeping every layer.
Measured from the group's start, an occurrence running from $t_{1}$ to $t_{2}$ occupies positions $\left[ \left\lfloor t_{1}/r \right\rfloor, \left\lceil t_{2}/r \right\rceil \right)$, or the first position when that range is empty.
We average the hidden states over that range.

For the text stream we write the same group as an \gls{ipa} string between slashes, as broad transcription conventionally is written and as dictionaries give it, and pass it through the same decoder.
A Sanzhi Dargwa phrase reads \trans{\textipa{/muX:raj daxul/}}.
We locate each occurrence by the character(s) it occupies, then average over every token whose span overlaps them, since a token may cover several symbols or only part of one.

\subsection{Feature directions from minimal pairs}
\label{sec:directions}

Each phoneme is described by a vector of binary distinctive features, read from a feature table \cite{mortensenPanPhonResourceMapping2016}.
A minimal pair is two phoneme types whose vectors differ in exactly one position.
We always subtract the member $x^{-}_{i}$ that lacks the feature from the member $x^{+}_{i}$ that has it, so that every pair of a feature is oriented alike.

Write $v(\cdot)$ for a phoneme type's mean representation at one layer.
For pair $\{x^{+}_{i}, x^{-}_{i}\}$ of a feature with $n$ pairs, the offset is $d_{i} = v(x^{+}_{i}) - v(x^{-}_{i})$ and its unit vector is $\hat{d}_{i}$.
Subtraction removes whatever the two members share, including the mean of the whole space, which is large in these anisotropic representations \cite{ethayarajhHowContextualAre2019,timkeyAllBarkNo2021}.

Their agreement $A$ is the mean cosine between those unit offsets,
\begin{equation}
  A = \frac{2}{n(n-1)} \sum_{i<j} \langle \hat{d}_{i}, \hat{d}_{j} \rangle .
  \label{eq:agreement}
\end{equation}
$A=1$ means every pair points the same way.
The feature's direction is $u = \bar{d} / \lVert \bar{d} \rVert$ for $\bar{d} = \frac{1}{n} \sum_{i} \hat{d}_{i}$.
We compute $A$ and $u$ separately for the audio and the text stream.

Our primary measure is the cosine similarity between a feature's directions in the two streams,
\begin{equation}
  c = \langle u_{\mathrm{audio}}, u_{\mathrm{text}} \rangle .
  \label{eq:cosine}
\end{equation}
Both are taken at the same layer, and no alignment step is needed, since both directions lie in the same decoder's space.

\subsection{What each measure is compared against}
\label{sec:references}

For arbitrary pairs, the distribution of $A$ is wider when there are fewer pairs, and that of $c$ is not centred on zero, because one pair's offsets in audio and in text both reflect which phonemes the pair contains.
Neither distribution can be computed, so we build a reference by pairing at random, in the manner of a control task \cite{hewittDesigningInterpretingProbes2019}.
For each of $B = 2000$ repetitions we draw, uniformly and without replacement, as many pairs of distinct phoneme types as the feature itself has, and recompute the quantities above.
Each pair's members are taken in the order drawn, so its orientation, like the pairing, does not come from a feature.
Within a repetition the same pairing is used on both streams, since pairing them independently would compare directions built from different phonemes and lower the reference.
With $k$ of the $B$ repetitions reaching the observed value, we report $p = (k+1)/(B+1)$ \cite{phipsonPermutationPvaluesShould2010}.

We also take the cosine of a feature's audio direction with each other feature's text direction.
The largest of those is its cross-feature value, and comparing $c$ against it asks whether the agreement is specific to that feature.

\subsection{Testing across languages}
\label{sec:across}

The language is our unit of replication.
For a feature we take the median $c$ over languages at each layer, then the largest of those medians.
The layer is not fixed in advance, since the depth at which the two streams agree most differs between models.

The reference therefore has to account for that choice.
The $b$th repetition from \cref{sec:references} uses its own pairing in every language and keeps it at every layer.
We then take the median over languages at each layer, and the largest of those medians.
The repetitions are independent across languages, so how they are paired does not matter.
We test every model against every feature, so we report a Benjamini--Hochberg $q$ \cite{benjaminiControllingFalseDiscovery1995} across those 42 tests alongside $p$.
The fractions in \cref{tab:structure} describe where a comparison holds across depth rather than adding hypotheses, so we leave their $p$ uncorrected.

We also ask whether a feature is one direction in a model or one per language, since a direction separating voiced from voiceless phonemes in one language might reflect that inventory rather than voicing in general.
For a feature and a stream, the across-language agreement is the mean cosine between the languages' directions, which is \cref{eq:agreement} applied across languages rather than across the pairs of one feature.
Its reference shuffles, within each language, which feature each of that language's directions belongs to.
Every language therefore keeps the directions it had, and only the correspondence of features between languages is destroyed.

\section{Results}
\label{sec:results}

\subsection{Features with matching directions in audio and in text}
\label{sec:results-cross-modal}

\begin{figure*}[t]
  \centering
  \includegraphics[width=\textwidth]{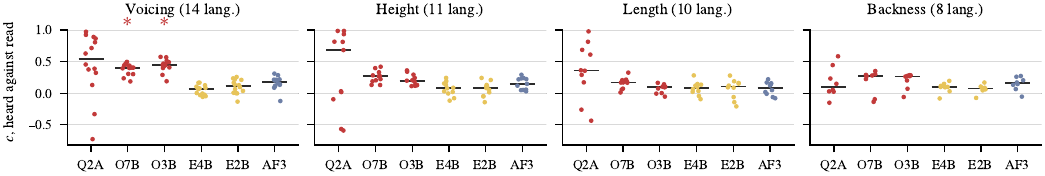}
  \caption{$c$ for each language, at the layer where the median over languages is largest.
  Bars mark the median and an asterisk marks $q \leq 0.05$. Models are Qwen2-Audio 7B (Q2A),
  Qwen2.5-Omni 7B and 3B (O7B, O3B), Gemma 4 E4B and E2B, and Audio-Flamingo 3 (AF3).}
  \label{fig:per-language}
\end{figure*}

Voicing has matching directions in the two Qwen2.5-Omni models, at $+0.40$ and $+0.45$ with $q = 0.011$ (\cref{tab:cross-modal}).
All 14 languages are positive in both (\cref{fig:per-language}), and every language also exceeds its cross-feature value.
No other combination of model and feature reaches $q \leq 0.05$.
Four of the 42 combinations reach $p \leq 0.05$ against 2.1 expected, and the other two have only three and four languages, exceeding their reference by $0.03$ or less.

Qwen2-Audio's voicing median is the largest in the table at $+0.54$, and its reference is $+0.75$.
Its two streams already agree about arbitrary phoneme pairs more than they agree about voicing.
The reference differs more between models than any feature effect does, from $+0.11$ in Gemma 4 E4B to $+0.75$ in Qwen2-Audio.
No median exceeds its own reference by more than $0.10$.

The reference and the cross-feature value in \cref{tab:cross-modal} answer different questions.
Qwen2.5-Omni-7B has all 11 languages positive for height and all 11 exceeding their cross-feature value, yet its $+0.27$ falls below a reference of $+0.36$.
A feature can be the best-matching one and still be unremarkable against arbitrary phoneme pairs.

\subsection{Which stream represents the features better}
\label{sec:results-modality}

\begin{table}[t]
\centering
\small
\setlength{\tabcolsep}{2.5pt}
\caption{Fraction of a model's layers at which each comparison holds, taking $p \leq 0.05$ as the criterion. \emph{Wthn}: A stream's own offsets agree more than arbitrary pairings of the same phonemes, median over languages. \emph{Acrs}: The languages' directions agree more with each other than with other features' directions. In each pair of rows the larger of the two values is set in bold.}
\label{tab:structure}
\begin{tabular}{@{}llcccccc@{}}
\hline
 & & \multicolumn{2}{c}{Voicing} & \multicolumn{2}{c}{Height} & \multicolumn{2}{c}{Length} \\
Model & Stream & Wthn & Acrs & Wthn & Acrs & Wthn & Acrs \\
\hline
Qwen2-Audio 7B & heard & \textbf{0.09} & \textbf{0.30} & 0.00 & 0.00 & 0.21 & 0.97 \\
 & read & 0.00 & 0.21 & \textbf{0.09} & \textbf{0.06} & \textbf{0.70} & \textbf{1.00} \\
\cline{2-8}
Qwen2.5-Omni 7B & heard & \textbf{0.84} & \textbf{1.00} & \textbf{0.52} & \textbf{1.00} & \textbf{1.00} & 0.83 \\
 & read & 0.00 & 0.17 & 0.14 & 0.14 & 0.19 & \textbf{1.00} \\
\cline{2-8}
Qwen2.5-Omni 3B & heard & \textbf{0.85} & \textbf{1.00} & \textbf{0.51} & \textbf{1.00} & \textbf{1.00} & 0.97 \\
 & read & 0.00 & 0.24 & 0.22 & 0.11 & 0.26 & \textbf{1.00} \\
\hline
Gemma 4 E4B & heard & 0.00 & 0.02 & 0.00 & 0.00 & 0.08 & 0.51 \\
 & read & \textbf{0.05} & \textbf{0.84} & \textbf{0.42} & \textbf{0.42} & \textbf{1.00} & \textbf{1.00} \\
\cline{2-8}
Gemma 4 E2B & heard & 0.00 & 0.03 & 0.00 & 0.00 & 0.01 & 0.83 \\
 & read & \textbf{0.07} & \textbf{0.83} & \textbf{0.39} & \textbf{0.33} & \textbf{1.00} & \textbf{1.00} \\
\hline
Audio-Flamingo 3 & heard & 0.00 & \textbf{1.00} & 0.03 & 0.07 & \textbf{0.98} & 0.97 \\
 & read & 0.00 & 0.17 & \textbf{0.14} & \textbf{0.14} & 0.17 & \textbf{1.00} \\
\hline
\end{tabular}
\end{table}

The models fall into two patterns (\cref{tab:structure}).
A stream counts as representing a feature at a layer when $A$ there exceeds its reference, and the \emph{Wthn} columns give the median over languages of that count, as a fraction of the model's depth.
The two Qwen2.5-Omni models represent voicing when they hear it and not when they read it, at 0.84 and 0.85 of their layers versus no layers.
The Gemma models show the reverse for height and length, representing them at 0.39 to 1.00 of their layers when reading and at 0.00 to 0.08 when hearing.

Each family includes two model sizes, so the family and not the size predicts which pattern a model shows.
The Qwen models initialise their audio encoder from Whisper \cite{radfordRobustSpeechRecognition2023} and train it against the decoder \cite{chuQwen2AudioTechnicalReport2024,xuQwen25OmniTechnicalReport2025}, whereas Gemma uses a Conformer encoder that stays frozen throughout pre-training \cite{teamGemma4Technical2026}.
With two families we cannot separate the effect of the encoder's architecture from whether it was trained.

Model size does not predict the median $c$ either.
For voicing the smaller model of each pair has the larger median $c$, $+0.45$ against $+0.40$ in Qwen2.5-Omni and $+0.11$ against $+0.07$ in Gemma, while for height and length the larger Qwen2.5-Omni model has the larger median (\cref{tab:cross-modal}).

The \emph{Acrs} columns ask something else: whether a model uses one direction for a feature in every language, or a separate direction in each.
A single value exists per layer here, so those columns give the fraction of layers directly rather than a median over languages.
The two questions have different answers.
Gemma 4 E2B represents length when it hears it at only 0.01 of its layers, yet at 0.83 of them the languages agree on one length direction.
A language's direction for a feature is the mean of several offsets, and a mean can point consistently even when the offsets it averages disagree with each other.
Representing a feature within a single language is therefore not a precondition for sharing a direction across languages.

\subsection{One direction across languages, or one per language}
\label{sec:results-across}

In three of the six models the languages' voicing directions in audio agree more than the reference at every layer (\cref{tab:structure}).
They agree at $+0.54$ and $+0.57$ in the two Qwen2.5-Omni models and at $+0.52$ in Audio-Flamingo 3, with that reference within $0.01$ of zero.
In the two Qwen2.5-Omni models every one of the 91 language pairs agrees, the weakest at $+0.31$, whereas two of Audio-Flamingo 3's pairs point opposite ways.
Qwen2-Audio reaches $+0.49$, but at only 10 of its 33 layers, and the two Gemma models reach $+0.19$ and $+0.15$ at one layer each.

In Audio-Flamingo 3 the median language has no layer at which its own voicing offsets exceed their reference, yet the languages agree with each other at every layer.
Across all models and features, a feature's agreement within a language and its across-language agreement correlate at $+0.67$ in audio and $+0.73$ in text.
A model that represents a feature within languages usually shares a direction across them as well.

Length in text has the highest across-language agreement of any feature, $+0.88$ or higher in every model, and it peaks at the first or second layer in all six.
Length is the one feature represented by adding the length mark (\textipa{:}) rather than by changing a symbol, so the agreement is present in the token embeddings rather than built by the decoder.
A shared written mark can give a trivially shared direction, without any phonological knowledge.

\section{Discussion}
\label{sec:discussion}

A feature's offset uses every occurrence of one phoneme against every occurrence of another, so any systematic difference in where the two occur enters the offset, and the design cannot remove that.
We measure that difference as the distance between the two members' distributions over neighbouring phonemes, and its correlation with $c$ is $-0.01$ for voicing, $+0.05$ for height and $+0.06$ for length.
At the embedding layer $c$ never exceeds $+0.10$ for any model or feature, so the voicing agreement in Qwen2.5-Omni is built by the decoder rather than inherited from the tokeniser.
Splitting every phoneme's occurrences in two gives two estimates of each direction, agreeing at a median of $0.82$ in audio and $0.96$ in text, so most estimates are well determined even though the weakest are not.

A model that mapped audio into a rotated copy of its text space would preserve every distance between phonemes while giving a generic direction zero cosine with its counterpart.
That is not what we observe for voicing in the two Qwen2.5-Omni models, though we cannot rule out a rotation that leaves some directions fixed.
A $c$ near zero elsewhere need not mean the two streams share no structure at all.
We check with \gls{rsa} \cite{kriegeskorteRepresentationalSimilarityAnalysis2008}, correlating the two streams' phoneme-similarity matrices by rank against a reference that shuffles which phoneme corresponds to which.
This comparison needs no minimal pairs, so all 15 languages contribute, and every model has a layer at which they all exceed that reference.

Both directions in $c$ are taken at the same layer, and in 12 of the 18 combinations in \cref{tab:structure} no layer has a majority of languages exceeding the reference in both streams.
A feature encoded early in audio and late in text would therefore read as absent, and ruling that out needs a search over pairs of layers.

After correction, agreement between the two streams holds only for voicing and only in the two Qwen2.5-Omni models, where comparing against zero would have found it in 15 of the 42 combinations instead.
Agreement across languages is wider.
In three of the six models the languages agree on one voicing direction in audio.

\bibliographystyle{IEEEbib}
\bibliography{library,doreco}

\end{document}